\documentclass[a4paper]{article}

\usepackage{INTERSPEECH2021}
\usepackage{todonotes}
\usepackage{adjustbox}
\usepackage{comment}
\usepackage{adjustbox}
\usepackage{multirow}
\usepackage{hyperref}
\usepackage[normalem]{ulem}

\def\JNdel#1{\bgroup\markoverwith{\textcolor{blue}{\rule[0.5ex]{2pt}{1pt}}}\ULon{#1}}

\def\ABdel#1{\bgroup\markoverwith{\textcolor{green}{\rule[0.5ex]{2pt}{1pt}}}\ULon{#1}}

\title{Comparing Acoustic-based Approaches for Alzheimer's Disease Detection}
\name{Aparna Balagopalan$^{1,2}$, Jekaterina Novikova$^1$}
\address{
  $^1$Winterlight Labs, Toronto, Canada \\
  $^2$University of Toronto, Toronto, Canada}
\email{aparna@winterlightlabs.com, jekaterina@winterlightlabs.com}

\begin{document}

\maketitle
\begin{abstract}
Robust strategies for Alzheimer's disease (AD) detection are important, given the high prevalence of AD. 
In this paper, we study the performance and generalizability of three approaches for AD detection from speech on the recent ADReSSo challenge dataset:
1) using conventional acoustic features 2) using novel pre-trained acoustic embeddings 3) combining acoustic features and embeddings. We find that while feature-based approaches have a higher precision
, classification approaches relying on pre-trained embeddings prove to have a higher, and more balanced cross-validated performance across multiple metrics of performance. Further, embedding-only approaches are more generalizable. Our best model outperforms the acoustic baseline in the challenge by 2.8\%.
\end{abstract}

\section{Introduction}

Alzheimer's disease (AD) is an irreversible, progressive brain disorder that slowly destroys memory and thinking skills. Research into the early assessment of Alzheimer's dementia is becoming increasingly more important, as the proportion of people affected by it grows every year~\cite{cahill2020whoplan}. Changes in cognitive ability due to neurodegeneration associated with AD lead to a progressive decline in memory and language quality~\cite{reitz2011epidemiology}.

Studies have shown that valuable clinical information indicative of cognition can be obtained from spontaneous speech elicited using pictures~\cite{borod1980normative}. Studies have capitalized on this clinical observation, using speech analysis, natural language processing (NLP), and machine learning (ML) to distinguish between speech from healthy and cognitively impaired participants in datasets of semi-structured speech tasks such as picture description \cite{fraser2016linguistic,balagopalan2018effect,zhu2019detecting}. As such, this approach shows a potential to serve as quick, objective, and non-invasive  assessments  of  an  individual's  cognitive  status. The ADReSSo Challenge \cite{luz2021detecting} aims to generate systematic evidence for these promises towards their clinical implementation. 

ADReSSo (Alzheimer's Dementia Recognition through Spontaneous Speech \textit{only}) Challenge is a shared task for the systematic comparison of approaches to the detection of cognitive impairment and decline based on spontaneous speech. In 2021, it focuses mainly on acoustic characteristics of speech, requiring the creation of models straight from speech, without manual transcription. 

In  this  work,  we  develop  ML  models  to  detect  AD  from speech using picture description data with the demographically-matched ADReSSo challenge speech dataset. Following the previous work on comparing the linguistic approaches to AD detection from speech~\cite{balagopalan2020bert}, we compare the following \textit{acoustic}-based approaches to detect AD:

\begin{enumerate}
\item \textbf{Extracting conventional acoustic features from speech:} with this approach, we extract acoustic features from the audio files for binary AD vs non-AD classification. The features extracted are informed by previous clinical and ML research in the space of cognitive impairment detection~\cite{fraser2016linguistic,alonso2015new}.

\item \textbf{Using pre-trained deep neural models:} with this approach, we embed the raw audio into representations using the last hidden state of the pre-trained wav2vec 2.0 model~\cite{baevski2020wav2vec} and classify audio samples using these embeddings as input to simple classifiers.

\item \textbf{A combination of both approaches.} Here, we combine the two approaches and make use of both engineered fesatures and audio representations generated using a pre-trained deep neural model. 
\end{enumerate}

In this paper,  we evaluate performance of the three approaches on both the ADReSSo train dataset, and on the unseen test set. We find that models based on conventional acoustic features are best suited for the AD screening tasks, as they are able to achieve very high precision, while under performing on the rest of evaluation metrics, such as accuracy, recall and F1 score. For the cases when a more balanced performance is required, a combination of conventional acoustic features and pre-trained deep neural models or pre-trained deep neural models alone are more promising methods, as they achieve higher performance and generalize well to unseen data.

The main contributions of our paper are as follows:
\begin{itemize}
\item We use audio representations extracted from the pre-trained deep neural model wav2vec 2.0~\cite{baevski2020wav2vec} for the task of AD classification from speech. To the best of our knowledge, this method and this model were never applied before for such a classification task. This model  outperforms the ADReSSo baseline model by 2.82\%. 
\item We present the model that combines conventional acoustic features with pre-trained representations of speech.
\item We carefully compare three different acoustic-based approaches for AD detection, which allows us to draw
more general conclusions on performance and generalizability of acoustic models.
\end{itemize}

\section{Background}

\subsection{Extracting conventional acoustic features from speech}

Some of the conventional features employed to describe acoustic characteristics of the voice applied to AD detection, include fundamental frequency, jitter and shimmer~\cite{pulido2020alzheimer}. In addition to these features, there is also a range of elements properly accompanying linguistic emissions and which constitute signs and clues but are not verbal. These characteristics of speech are called paralinguistic features and have been used to obtain information from the patient by means of the statistics of e.g. Cepstral Mel-Frequency Components (MFCC),
among others~\cite{alonso2015new}. Different combinations of these features were used in multiple previous studies when detecting AD from the speech collected via picture description tasks~\cite{balagopalan2018effect,zhu2019detecting,zhu2018semi,balagopalan2020impact}. These works have provided clear evidence on the potential of using simple spoken tasks and conventional acoustic features to automatically assess early dementia and its progression as well as to demonstrate that technology allows automatic detection of AD.


\subsection{Using pre-trained deep neural methods}

In the recent years, pre-training of deep neural networks has emerged as an effective approach to overcome the problem of data scarcity~\cite{young2018recent,devlin2019bert,schneider2019wav2vec,baevski2020wav2vec}. The key idea of such a technique, which is also called ``transfer learning", is to learn general representations in a setup where substantial amounts of labeled or unlabeled data is available and to leverage the learned robust representations to improve performance on a downstream task for which the amount of data is limited. 

In natural language processing, one of the most popular transfer learning models is BERT~\cite{devlin2019bert}, which trains ``contextual embeddings" wherein a representation of a sentence (or transcript) is influenced by the context in which the words occur in sentences. In the field of speech processing, transfer learning models are mainly used for the purpose of automatic speech recognition~\cite{baevski2020wav2vec}. We focus on using pre-trained embeddings from a self-supervised audio representation model, wav2vec 2.0 ~\cite{baevski2020wav2vec}, for the task of AD detection from speech. In the wav2vec 2.0 model, audio is encoded via a multi-layer convolutional neural network and then masks spans of the resulting latent speech representations similar to masked language modeling~\cite{devlin2019bert}. The latent representations are fed to a Transformer network~\cite{vaswani2017attention} to build contextualized representations.


For the task of AD detection, transfer learning is particularly useful, as it is difficult and costly to collect labelled data. While several prior works have used pre-trained acoustic embeddings such as x-vectors and i-vectors for AD detection from speech~\cite{pompiliinesc,hauptman2019identifying}, to the best of our knowledge, no works have utilized self-supervised representations of speech for AD detection. Hence, we aim to benchmark a self-supervised representation learning methodology for AD detection.


\subsection{A combination of pre-trained and conventional approaches}
Incorporating domain-specific external knowledge in neural language representations is a field of  research  that  has  been  actively  explored with both acoustic and linguistic embeddings~\cite{balagopalan2020augmenting,cai2019multi}. However, a large amount of prior work is either focused on linguistic based approaches or using simple acoustic embeddings such as x-vectors~\cite{baevski2020wav2vec}. In contrast, we are using a combination of state-of-the art self-supervised techniques for speech representations and combining these with domain-knowledge informed conventional speech features.

\section{Methodology}

\subsection{Dataset}
The ADReSSo dataset we use in this work consists of set of speech recordings of picture descriptions produced by cognitively normal (or healthy) subjects and patients with an AD diagnosis, who were asked to describe the Cookie Theft picture~\cite{goodglass1983boston} from the Boston Diagnostic Aphasia Examination~\cite{goodglass1983boston}. There are speech samples from 237 participants in total, out of which 166 are in the training set, and 71 in the test set ($70/30$ split balanced for demographics). Out of the samples in the training set, 83 were cognitively healthy, while 83 had AD. The prediction dataset used is matched for age and gender so as to minimise risk of bias in the prediction tasks using a propensity score.
Along with the recorded speech data, the dataset also included
segmentation profiles for optional use. For all our acoustic approaches, we use the full unsegmented audio, following the baseline acoustic approach~\cite{luz2021detecting} and prior works~\cite{pompiliinesc,sarawgimultimodal}.

\subsection{Feature Extraction}

\subsubsection{Using conventional acoustic features from speech}
\label{sec:features-eng}
We extract 168 acoustic features from the unsegmented speech audio files. Those include several statistics such as mean, variance, kurtosis, etc. of mel-frequency cepstral coefficients (MFCCs), following prior work~\cite{fraser2016linguistic}. We only use samples from the audio component provided with the dataset, and do not perform any Automatic Speech Recognition or linguistic analyses.

\subsubsection{Using embeddings from pre-trained deep neural models for audio representation}
\label{sec:features-embed}

In order to create audio representations using this approach, we make use of the huggingface\footnote{https://huggingface.co/models} implementation of the wav2vec~2.0~\cite{baevski2020wav2vec} base model \textit{wav2vec2-base-960h}. This base model is pretrained and fine-tuned on 960 hours of Librispeech on 16kHz sampled speech audio. We first represent each unsegmented audio file as a waveform with librosa\footnote{https://librosa.org/}. We then tokenize waveforms using \textit{Wav2Vec2Tokenizer} and if necessary, divide them into smaller chunks (with the maximum size of 320000 in our case) to fit into memory, afterwards we feed them into the wav2vec~2.0 model. The last hidden state of the model is used as an embedded representation of audio. When tokenized waveforms are divided into several chunks, the mean value of all chunks
is computed to generate the final embedding.

\subsubsection{A combination of both approaches}
\label{sec:features-combo}
In this approach, we study if engineered feature and pre-trained embeddings can augment each other by concatenating the representations at audio sample level. We combine representations from pre-trained models and conventional acoustic features by simply concatenating them. Hence, the dimension of the overall representation is the sum of the individual dimensions (936-dimensional representation vector overall). 


\begin{table*}[ht!]
\caption{10-fold CV and LOSO-CV results averaged across all the folds on the ADReSSo train set. Bold indicates the best performing approach for each model, bold+italics indicate the best overall performance for the metric.}
\label{tab:train-results}
\begin{adjustbox}{max width=0.9\textwidth, center}
\begin{tabular}{lrrrrrrrr}
 & \multicolumn{2}{c}{\textbf{Accuracy}} & \multicolumn{2}{c}{\textbf{Precision}} & \multicolumn{2}{c}{\textbf{Recall}} & \multicolumn{2}{c}{\textbf{F1}} \\
\textbf{Model} & \multicolumn{1}{c}{\textbf{10-fold CV}} & \multicolumn{1}{c}{\textbf{LOSO CV}} & \multicolumn{1}{c}{\textbf{10-fold CV}} & \multicolumn{1}{c}{\textbf{LOSO CV}} & \multicolumn{1}{c}{\textbf{10-fold CV}} & \multicolumn{1}{c}{\textbf{LOSO CV}} & \multicolumn{1}{c}{\textbf{10-fold CV}} & \multicolumn{1}{c}{\textbf{LOSO CV}} \\
\hline \hline
LR-feat & 0.6084 & 0.6386 & 0.6774 & \textbf{0.7213} & 0.4828 & 0.5057 & 0.5638 & 0.5946 \\
LR-embed & \textbf{0.6867} & \textbf{0.6747} & 0.6882 & 0.6737 & \textit{\textbf{0.7356}} & \textit{\textbf{0.7356}} & \textit{\textbf{0.7111}} & \textbf{0.7033} \\
LR-combo & 0.6807 & 0.6687 & \textbf{0.7024} & 0.6951 & 0.6782 & 0.6552 & 0.6901 & 0.6746 \\
\hline
SVM-feat & 0.6265 & 0.6566 & \textit{\textbf{0.8571}} & \textit{\textbf{0.9167}} & 0.3448 & 0.3793 & 0.4918 & 0.5366 \\
SVM-embed & 0.6687 & 0.6566 & 0.6818 & 0.6705 & \textbf{0.6897} & \textbf{0.6782} & 0.6857 & \textbf{0.6743} \\
SVM-combo & \textit{\textbf{0.6928}} & \textbf{0.6807} & 0.7308 & 0.7297 & 0.6552 & 0.6207 & \textbf{0.6909} & 0.6708 \\
\hline
NN-feat & 0.6084 & 0.6265 & 0.6833 & 0.6923 & 0.4713 & 0.5172 & 0.5578 & 0.5921 \\
NN-embed & \textbf{0.6747} & 0.6566 & \textbf{0.6774} & 0.6705 & \textbf{0.7241} & 0.6782 & \textbf{0.7000} & 0.6743 \\
NN-combo & 0.6506 & \textit{\textbf{0.6928}} & 0.6706 & \textbf{0.7000} & 0.6552 & \textbf{0.7241} & 0.6628 & \textit{\textbf{0.7119}} \\
\hline
DT-feat & 0.5964 & 0.5843 & 0.6190 & 0.6071 & 0.5977 & 0.5862 & 0.6082 & 0.5965 \\
DT-embed & 0.6446 & 0.6807 & 0.6591 & 0.6809 & 0.6667 & \textit{\textbf{0.7356}} & 0.6629 & 0.7072 \\
DT-combo & \textbf{0.6506} & \textbf{0.6867} & \textbf{0.6593} & \textbf{0.6882} & \textbf{0.6897} & \textit{\textbf{0.7356}} & \textbf{0.6742} & \textbf{0.7111}
\end{tabular}
\end{adjustbox}
\end{table*}

\begin{table}[ht]
\caption{AD detection results on unseen, held-out ADReSS test set. Bold indicates the best result. The submissions attempted to the challenge are indicated in parentheses.}
\label{tab:test-results}
\begin{adjustbox}{max width=0.9\linewidth, center}
\begin{tabular}{lrrrr}
\textbf{Model} & \multicolumn{1}{c}{\textbf{Accuracy}} & \multicolumn{1}{c}{\textbf{Precision}} & \multicolumn{1}{c}{\textbf{Recall}} & \multicolumn{1}{c}{\textbf{F1}} \\
\hline \hline
Acoustic baseline & 0.6479 & \multicolumn{1}{l}{} & \multicolumn{1}{l}{} & \multicolumn{1}{l}{} \\
\hline
SVM-feat (Attempt-1) & 0.6479 & \textbf{0.9167} & 0.3143 & 0.4681 \\
LR-embed (Attempt-2) & 0.6056 & 0.6000 & 0.6000 & 0.6000 \\
DT-embed (Attempt-3) & 0.5775 & 0.5714 & 0.5714 & 0.5714 \\
SVM-embed (Attempt-4) & \textbf{0.6761} & 0.6364 & \textbf{0.8000} & \textbf{0.7089} \\
SVM-combo & 0.6338 & 0.6364 & 0.6000 & 0.6176 \\
NN-combo & 0.6620 & 0.6897 & 0.5714 & 0.6250 \\
\end{tabular}
\end{adjustbox}
\end{table}

\subsection{Evaluation Methods}
We evaluate performance primarily using accuracy scores, since all train/test sets are known to be balanced. We also report precision, recall, and F1 with respect to the positive class (AD).

\textbf{Cross-validation on ADReSSo train set:} We use two cross-validation strategies in our work – leave-one-subject-out (LOSO) CV and 10-fold CV. We report evaluation metrics with both these strategies for all models for direct comparison between approaches and with challenge baseline. 

\textbf{Predictions on ADReSSo test set:} We generate predictions from and focus our analyses on the top-performing classifiers -- where these classifiers are selected on the basis of highest LOSO-CV train accuracy, and challenge test predictions are generated from models trained on the last LOSO train split. We report performance on the challenge test set, using test set labels provided by the challenge organizers.

\subsection{Experiments}
\subsubsection{Using conventional acoustic features from speech}
We classify acoustic features (see Section~\ref{sec:features-eng}) extracted at sample-level with several conventional linear and non-linear ML models : Logistic regression (LR-feat), Support Vector Machines (SVM-feat), Neural Network (NN-feat), and Decision Tree (DT-feat). We  perform feature selection by choosing top-$k$ number of features, based on ANOVA F-value between label/features, where $k$ was set to $10$ based on LOSO-CV. All model hyper-parameters were set to their default values as on the scikit-learn~\cite{pedregosa2011scikit} implementation for each of these: logistic regression is trained with L2-penalty, SVM-feat is trained with a radial basis function kernel with kernel coefficient $0.001$, and regularization parameter set to $1$,  NN-feat used has 1 layer with 10 units, and DT-feat with minimum samples per split set to $2$.

\subsubsection{Using embeddings from pre-trained deep neural models for audio representation}
In order to leverage information encoded by pre-trained audio representation models, we extract embeddings from a pre-trained audio model, wav2vec 2.0, as a representation for each audio sample (see Section~\ref{sec:features-embed}). We then classify these by training: 
Logistic regression (LR-embed), Support Vector Machines (SVM-embed), Neural Network (NN-embed), and Decision Tree (DT-embed) with default hyper-parameters.

\subsubsection{Combining conventional features and embeddings from pre-trained deep neural models}
 Hence, a single representation with length equal to the sum of features and the embedding-dimension is obtained (see Section~\ref{sec:features-combo}). We then classify these by training the following models on these concatenated representations : Logistic regression (LR-combo), Support Vector Machines (SVM-combo), Neural Network (NN-combo), and Decision Tree (DT-combo). All model hyper-parameters were set to their default values as on the scikit-learn implementation for each of these (i.e., same as in previous sections).

\section{Results}

\subsection{AD detection results on the ADReSSo train set}

The results of both LOSO and 10-fold CV evaluation of classification models' performance show that the conventional features-based approach consistently under performs in terms of accuracy, recall and F1 score (see Table~\ref{tab:train-results}). The feature-based approach, however, outperforms all the other approaches in terms of precision with the SVM model. With the NN and DT models, feature-based approach reaches high precision levels, although is not the best performing 
NN models (see Table~\ref{tab:train-results}). 

Embedding-based and combo approaches compete for best performance in terms of accuracy, recall and F1 score. With logistic regression, embedding-based approach outperforms all the other approaches both in terms of accuracy, recall and F1. With the neural net model, the type of cross-validation determines whether it is the embedding-based or the combo approach that performs the best. With decision trees, however, the combination of conventional features and embeddings of speech perform the best in all the cases, even in terms of precision. 

\subsection{AD detection results on the unseen ADReSSo test set}

\begin{table*}[ht!]
\caption{Difference between test performance and performance of performance of the model evaluated using 10-fold CV (i.e. \textit{w/ 10-fold}) and LOSO CV (i.e. \textit{w/ LOSO}). Positive results indicate that test performance was higher. Bold indicates the model that outperforms the acoustic baseline.}
\label{tab:diff-test}
\begin{adjustbox}{max width=0.9\textwidth, center}
\begin{tabular}{lcccccccc}
 & \multicolumn{2}{c}{\textbf{Accuracy}} & \multicolumn{2}{c}{\textbf{Precision}} & \multicolumn{2}{c}{\textbf{Recall}} & \multicolumn{2}{c}{\textbf{F1}} \\
\textbf{Model} & \textbf{w/ 10-fold} & \textbf{w/ LOSO} & \textbf{w/ 10-fold} & \textbf{w/ LOSO} & \textbf{w/ 10-fold} & \textbf{w/ LOSO} & \textbf{w/ 10-fold} & \textbf{w/ LOSO} \\
\hline \hline
SVM-feat & \multicolumn{1}{l}{2.14\%} & -0.87\% & \multicolumn{1}{l}{5.96\%} & 0.00\% & -3.05\% & -6.50\% & -2.37\% & -6.85\% \\
LR-embed & -8.11\% & -6.91\% & -8.82\% & -7.37\% & -13.56\% & -13.56\% & -11.11\% & -10.33\% \\
DT-embed & -6.71\% & -10.32\% & -8.77\% & -10.95\% & -9.53\% & -16.42\% & -9.15\% & -13.58\% \\
\textbf{SVM-embed} & +0.74\% & +1.95\% & -4.54\% & -3.41\% & +11.03\% & +12.18\% & +2.32\% & +3.46\%\\
SVM-combo & -5.90\%& -4.69\% & -9.44\% & -9.33\% & -5.52\% & -2.07\% & -7.33\% & -5.32\%\\ 
NN-combo & +1.14\%& -3.08\% & +1.91\% & -1.03\% & -8.38\% & -15.27\% & -3.78\% & -8.69\%\\ 
\end{tabular}
\end{adjustbox}
\end{table*}

\begin{table}[t]

\caption{Difference between LOSO and 10-fold CV classification performance. Positive results indicate that LOSO CV performance was higher than that of 10-fold CV.}
\label{tab:diff-loso-10fold}
\begin{adjustbox}{max width=0.9\linewidth, center}
\begin{tabular}{lllll}
\textbf{Model} & \multicolumn{1}{c}{\textbf{Accuracy}} & \multicolumn{1}{c}{\textbf{Precision}} & \multicolumn{1}{c}{\textbf{Recall}} & \multicolumn{1}{c}{\textbf{F1}} \\
\hline \hline
LR-feat & 3.02\% & 4.39\% & 2.29\% & 3.08\% \\
SVM-feat & 3.01\% & 5.96\% & 3.45\% & 4.48\% \\
NN-feat & 1.81\% & 0.90\% & 4.59\% & 3.43\% \\
DT-feat & \multicolumn{1}{r}{-1.21\%} & \multicolumn{1}{r}{-1.19\%} & \multicolumn{1}{r}{-1.15\%} & \multicolumn{1}{r}{-1.17\%} \\
\hline
LR-embed & \multicolumn{1}{r}{-1.20\%} & \multicolumn{1}{r}{-1.45\%} & \multicolumn{1}{r}{0.00\%} & \multicolumn{1}{r}{-0.78\%} \\
SVM-embed & \multicolumn{1}{r}{-1.21\%} & \multicolumn{1}{r}{-1.13\%} & \multicolumn{1}{r}{-1.15\%} & \multicolumn{1}{r}{-1.14\%} \\
NN-embed & \multicolumn{1}{r}{-1.81\%} & \multicolumn{1}{r}{-0.69\%} & \multicolumn{1}{r}{-4.59\%} & \multicolumn{1}{r}{-2.57\%} \\
DT-embed & 3.61\% & 2.18\% & 6.89\% & 4.43\% \\
\hline
LR-combo & \multicolumn{1}{r}{-1.20\%} & \multicolumn{1}{r}{-0.73\%} & \multicolumn{1}{r}{-2.30\%} & \multicolumn{1}{r}{-1.55\%} \\
SVM-combo & \multicolumn{1}{r}{-1.21\%} & \multicolumn{1}{r}{-0.11\%} & \multicolumn{1}{r}{-3.45\%} & \multicolumn{1}{r}{-2.01\%} \\
NN-combo & 4.22\% & 2.94\% & 6.89\% & 4.91\% \\
DT-combo & 3.61\% & 2.89\% & 4.59\% & 3.69\%
\end{tabular}
\end{adjustbox}
\end{table}

Models were selected based on their performance on the train set -- magnitude as well as well as stability of accuracy in the two modes of CV -- to submit to the ADReSSo challenge - SVM-feat to represent the feature-based approach, LR-embed, SVM-embed, and DT-embed to represent the embedding-based approach. SVM-combo and NN-combo represented the combined approach. Out of these models, SVM-feat had the highest precision (see Table~\ref{tab:test-results}~\footnote{Note: This table is corrected from the \href{https://www.isca-speech.org/archive/pdfs/interspeech_2021/balagopalan21_interspeech.pdf}{prior published version}. Please cite this version of the paper.}). Out of all the models tested on the test set, the SVM-embed model achieved the best performance in terms of accuracy, recall and F1, and was able to beat the baseline acoustic model by 2.82\%. 


\section{Discussion}

\subsection{Classification performance}

The models employing the feature-based approach are performing the best on the train data in terms of precision, when evaluated using LOSO CV. Both the highest achieved precision level (91.67\% with SVM-feat) and the second-best result (72.13\% with LR-feat) are achieved by the models that make use of conventional acoustic features. The test results reinforce the precision capability of the feature-based model, with the SVM-feat model achieving much higher precision on the unseen test set than any other model. However, feature-based models perform very poorly in terms of recall. As such, feature-based models could be a good candidate in a real-life deployment of AD screening models, when high precision is much more important than recall or overall accuracy.

There is no clear difference in train performance between the embedding-based and combo approaches. Combo models tend to perform better when evaluated using LOSO CV method, while embedding-based models often outperform combo models when evaluated using 10-fold CV. The test results show that the NN-combo and SVM-embed models outperform the baseline, but more detailed analysis of generalizability of the three approaches is necessary. We also note that NN-based models have high variance across runs. 


\subsection{Generalizability}
In order to assess generalizability of the models, we first calculate the difference in performance of a variety of six tested models on the test and the train sets (see Table~\ref{tab:diff-test}). 

Feature-based model (SVM-feat) has no gap between test and train performance in terms of precision, and even surpasses the 10-fold CV precision on the test set. This model also achieves the highest precision levels both on the train and test sets, which allows us to conclude that the feature-based approach is potentially the best candidate for models when precision is the most important metric. This is not surprising, as these acoustic features were extracted based on several prior works on the speech characterization of patients with AD~\cite{balagopalan2018effect,zhu2019detecting,zhu2018semi,fraser2016linguistic}. 


Most combination-based models have a substantial gap between the test performance and the cross-validated performance on the train set. The SVM-embed approaches on average, however, result in a smaller gap, especially in terms of accuracy, recall and F1 score. The SVM-embed model even performs better on the test set, when evaluated using recall and F1 metrics. 
Such a slight difference confirms that the embedding-based approach may be a more generalizable method to train the AD detection system, comparing to using the combination-based or feature-based representations of speech alone. However, statistical testing of performance on the test set and validation on more datasets are required to further ascertain this.

We also aim to understand whether 10-fold or LOSO CV is the best approach to evaluate the model in order to achieve the result that is close to the expected test performance. For this, we additionally compare performance of each model between LOSO and 10-fold CV (see Table~\ref{tab:diff-loso-10fold}). With the feature-based approach, LOSO CV results are usually higher than those of 10-fold CV. As these models are most suitable for precision-focused cases (as we discussed above) and given test precision does not differ much from the LOSO CV results, the best solution would be to use LOSO CV evaluation when training these models on the data similar in size to the ADReSSo dataset.

With embedding-based and combo models, 10-fold CV mostly tends to outperform LOSO CV. However, test performance is in most cases closer to the LOSO CV performance. So with these approaches, similarly as with the feature-based approach, relying on the LOSO CV would be a more reliable strategy, when data is similar to the ADReSSo dataset. Note that these findings are dependant on the speech representation and modelling strategies as well as dataset domain in our study.

\section{Conclusions and Future Work}
In this paper we study the performance of conventional acoustic feature-based  and pre-trained embedding-based classification approaches on Alzheimer's Disease detection from speech on the ADReSSo challenge dataset. We observe that feature-based approaches have a higher precision in general, and hence might be well-suited for screening AD via speech analysis. However, a more balanced performance across multiple-metrics is achieved by embedding-based approaches both while cross-validating on the train set and when tested on the unseen test set. Finally, we observe that a representation using pre-trained embeddings outperforms the challenge baseline, and attains an accuracy 2.8\% higher than the best acoustic baseline in the challenge. With our careful comparisons, we hope to contribute to principled evaluations of the performance and generalizability of classification strategies on the important task of Alzheimer's Disease detection from speech. 
In future work, we will focus on different strategies to combine and fine-tune embeddings and feature-based models.

\bibliographystyle{IEEEtran}

\bibliography{mybib}


\appendix
\end{document}